# Analysis of Weather and Time Features in Machine Learning-aided ERCOT Load Forecasting

Jonathan Yang
*Student Member, IEEE*
Department of Electrical and Computer Engineering
University of Houston
Houston, TX, USA
jjyang5@uh.edu

Mingjian Tuo
*Student Member, IEEE*
Department of Electrical and Computer Engineering
University of Houston
Houston, TX, USA
mtuo@uh.edu

Jin Lu
*Student Member, IEEE*
Department of Electrical and Computer Engineering
University of Houston
Houston, TX, USA
jlu27@cougarnet.uh.edu

Xingpeng Li
*Senior Member, IEEE*
Department of Electrical and Computer Engineering
University of Houston
Houston, TX, USA
xli82@uh.edu

*Abstract*— Accurate load forecasting is critical for efficient and reliable operations of the electric power system. A large part of electricity consumption is affected by weather conditions, making weather information an important determinant of electricity usage. Personal appliances and industry equipment also contribute significantly to electricity demand with temporal patterns, making time a useful factor to consider in load forecasting. This work develops several machine learning (ML) models that take various time and weather information as part of the input features to predict the short-term system-wide total load. Ablation studies were also performed to investigate and compare the impacts of different weather factors on the prediction accuracy. Actual load and historical weather data for the same region were processed and then used to train the ML models. It is interesting to observe that using all available features, each of which may be correlated to the load, is unlikely to achieve the best forecasting performance; features with redundancy may even decrease the inference capabilities of ML models. This indicates the importance of feature selection for ML models. Overall, case studies demonstrated the effectiveness of ML models trained with different weather and time input features for ERCOT load forecasting.

## I. Introduction

Electrical power systems are responsible for supplying energy to a large range of applications, ranging from consumer electronics such as personal computers to industrial appliances such as manufacturing equipment. The demand and importance for this commodity have continued to increase due to the fast growth of industrial electrification and consumer appliances such as electric vehicles [1]-[2]. This increasingly volatile use of electricity makes it more challenging to reliably operate the power grid. Much of the present electricity generation is still reliant on nonrenewable sources, and it is important to reduce losses from over-generator-commitment and prevent insufficient online capacity from under-generator-commitment by providing accurate load forecasting information to properly match electricity generation and electricity demand.

To support grid operations and enhance grid reliability, it is essential to be able to accurately predict the electrical load, such that the grid operators can make proper adjustments well in advance, pre-positioning the system to reliably supply all the load in a cost-effective manner. Accurate load forecasting can help minimize the risk of grid power unbalancing, power plant disconnection and load shedding [3]. Electric load forecasting can be divided into long-term load forecasting, medium-term load forecasting, and short-term load forecasting, with periods of years, months, and days/hours [4]-[5]. This work will focus on methods to improve the accuracy of intra-day short-term load forecasting.

Power demand is difficult to predict even in the short term since it fluctuates greatly and depends on various random factors, with the most important and volatile being human decisions [6]. Although multiple factors such as human decisions are unknown to grid operators, they may be predicted from known factors. A large portion of electricity consumption is used in some sort of temperature control: heating and cooling. It is projected that by 2050, 3.08 of 6.82 quads of electricity in the residential sector, and 1.69 of 5.45 quads in the commercial sector, will be used in relation to temperature control [7]. Time and weather conditions will greatly affect these temperature-dependent loads [8]. This indicates that fundamental time and weather conditions will also greatly impact the use of electricity. Therefore, known factors, such as temperature, that affect appliance needs can be used to predict power demand in a similar way as using the unknown factors, such as the status and on/off profiles of a large number of appliances.

As the advancement of machine learning (ML) technologies in recent years, the use of ML such as fully-connected neural network (FCNN) and convolutional neural network (CNN) has been growing significantly in many power system applications [9]-[12]. One major application is look-ahead load forecasting [13]-[17]; such ML-based load forecasting studies generally focus on the ML model performance in comparison to other benchmark methods. In [13], several load forecasting methodologies are presented, many of which use neural network models and draw comparisons with models created by other methods such as multiple regressions and fuzzy logic. In [14], feature selection was only briefly reviewed. In [15], nine categories of load forecasting methods are summarized; however, little explanation is given for the choice of selected features. The long short-term memory (LSTM) model that is designed to process time series data is adopted in [16] for aggregated residential load forecasting; however, the time and weather information that is important for load forecasting is not included as the input features. In [17], it also used load information as the only inputs to its developed deep learning model. Overall, the important weather and time information as well as the associated ablation studies are not considered when training the ML models in the aforementioned efforts.

Temporal features such as the time of the day and the day of the week may provide valuable information for load predictions as the electricity consumption is observed to have cycling patterns at different timescales [18]. Weather factors such as the wind speed may also have significant impacts on the load [19]-[20]. When used together, the individual contribution from each feature is masked by the presence of other features. However, it is also possible that the information from some features such as temperature is fully or mostly contained within one or multiple of the other features.

Therefore, to address the aforementioned gap, this paper conducts a study to identify features and feature combinations that provide the most information to load forecasting for deep learning methods. Various weather features are used independently and in combination with each other to train and compare different ML models. We examine various popular models including the support vector machine (SVM), FCNN, LSTM, and the long-term recurrent convolutional networks (LRCN) with variation to account for differences in model capabilities. SVM performs well with linear patterns and carefully selected features; LSTM is suited to capture temporal behavior of time-series inputs; and LRCN improves upon LSTM by better capturing spatial patterns. These ML models are tested on the Electric Reliability Council of Texas (ERCOT) system.

The remainder of this paper is organized as follows. Section II introduces the experimental process of the study along with a description of the tools that we used, namely structures of the neurons used in each of the neural networks. Section III presents the data that were used. Section IV provides a case study of the accuracy of various trained models compared by their inputs, model structure, and period of inputs. Finally, the concluding remarks and future work are provided in Section V.

## II. MACHINE LEARNING MODELS

### A. Support Vector Machine

SVM is a supervised ML algorithm that draws its foundations from the statistical learning theory. Thanks to its exceptional capacity for generalization, SVM has been found to be very effective in both classification and regression tasks particularly the tasks related to time series prediction.

The SVM will become support vector regressor (SVR) for regression tasks, which will use a cost function $f(x)$ to minimize the empirical risk [21]. The objective of SVM/SVR is to determine the parameters in function $f(x)$ that best models the training data and ensure it is as flat as possible. In the case of a linear cost function, it can be described as follows,

$$f(x) = \langle w, x \rangle + b \quad (1)$$

where $\langle \cdot, \cdot \rangle$ denotes the dot production. To obtain the optimal cost function, the SVM problem can be reformulated as a convex optimization problem as follows:

$$minimize \frac{1}{2} \|w^2\| \quad (2)$$

$$subject\ to \begin{cases} y_i - \langle w, x_i \rangle - b \leq \varepsilon \\ \langle w, x_i \rangle + b - y_i \leq \varepsilon \end{cases} \quad (3)$$

where $i$ is the training point index, and $\varepsilon$ is the error tolerance.

Constraint (3) ensures the absolute error is not greater than $\varepsilon$ for each training point. In this work, the SVM/SVR model is built based on the historical data in a look-back time window to subsequently predict the future load curves in a look-ahead time window. The SVM training is achieved by minimizing the sum of squares of error on the training set.

### B. Long Short-Term Memory

A recurrent neural network (RNN) represents an alternative deep learning framework that leverages its internal state to handle a sequence of inputs [9]. LSTM extends RNN, and it excels in capturing the temporal patterns present in time series input data. LSTM can learn intricate long-term relationships within sequential data such as load fluctuations. Fig. 1 illustrates an LSTM cell with gate control [16].

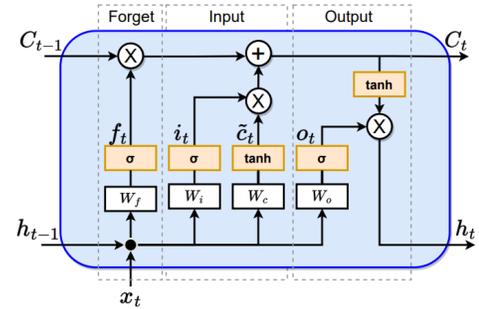

Fig. 1. Illustration of an LSTM cell [16].

The architecture of the LSTM model trained in this work is explained as follows. First, LSTM layers are used to handle the time-series data containing multi-dimensional inputs. The output of LSTM layers is flattened and then sent to fully-connected dense layer with dropout enabled. Lastly, a fully-connected dense layer is used to serve as the output layer. Each neuron corresponds to the predicted load of each hour in the look-ahead prediction window at one-hour resolution. The Adam with exponential decay is used as the optimizer to train the proposed LSTM model.

### C. Long-Term Recurrent Convolutional Network

LRCN utilizes both CNN layers and LSTM layers. It is able to harness the advancements in CNN as well as to capture temporal dependencies. LRCN has achieved success in various domains including computer vision and time-series analysis.

In the trained LRCN model, the first CNN layers are the first to process the input features. The outputs from CNN layers are then fed into the LSTM layers to extract the temporal features. Fully-connected dense layers are the downstream layers including the last output layer that produces the predicted loads. The basic architecture of the LRCN model is illustrated in Fig. 2; note that LRCN may also include dropout feature and require flatten to match dimension between two consecutive layers.

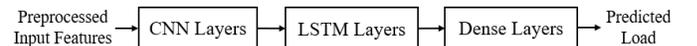

Fig. 2. Illustration of the LRCN model.

Another benchmark model implemented is FCNN that only consists of hidden dense layers. A sample FCNN model architecture can be illustrated by Fig. 2 after removing the two blocks representing CNN layers and LSTM layers respectively.

*D. Training and Testing Strategy*

Since different models and architectures may be better suited for different inputs, we trained and tested close to 200 ML models of varying layer organizations and feature combinations. We optimized various ML models including SVM, FCNN, LSTM, and LRCN, and used the best tuned models under each category to compare the prediction accuracy. For SVM, it could only produce a single-value output, which requires to separately train several models for each input feature combination, each one trained with all input features to predict one hour from the subsequent hours. We compared the models that have the best accuracies of each category, e.g., the most accurate FCNN model are compared to the most accurate LSTM.

To optimize the ML models, we tuned hyperparameters for the neural network layers within a limited range of neurons, adjusted the number and organization of neuron layers, and the number of training epochs. Changes to the network layers are determined by testing incremental adjustments and testing, while the number epochs are determined by the changes in training losses. In addition to experimentally optimizing ML model architectures for comparison between different ML models, the effects of different combinations of input features for the same model are also evaluated.

To evaluate a model's performance, the model is first trained on the true weather, time, and electrical load data obtained from the same region. The accuracy is obtained from testing on a dataset from the same source but distinct from the training and validation data, i.e. the dataset of the ERCOT region in a different period. For standard comparison of model accuracies, we used the measures of mean absolute percent error (MAPE) and the coefficient of determination (R2) as metrics of the difference between predictions from the actual loads. Our benchmark statistics used a prediction that simply took the most recent load and repeated that load as the estimate for the next $T2$ hours in the look-ahead prediction period.

It was intuitive to assume, which was later on proved to be incorrect, that by using all information available for training and testing an ML model, the model would produce the "best possible performance" of that design. Thus, when comparing models of different algorithms and architectures, we used all time and weather features to train and test the models. To optimize the performance of different model algorithms, we experimentally modified the model architecture through several iterations of training. We compared the model representing the "best possible performance" for each model type to determine the most optimal model type to use for load forecasting with weather and time features. To compare the effectiveness of including certain features in training a load forecasting model, we performed an ablation study by training LSTM models with certain features removed from ML inputs.

### III. DATA DESCRIPTION

*A. Data Sources*

Both historical load and weather data for a period of 2011-2021 are used in this work. The historical load profiles for the Texas grid at one-hour resolution are obtained from the ERCOT website [22]. The hourly weather data originally obtained from the Phase 2 of the North American Lan Data Assimilation System (NLDAS-2) [23] are cleaned up for the Texas footprints [24]. These weather data specifically include temperature (K), zonal wind speed (m/s), meridional wind speed (m/s), longwave radiation (LWRad) (W/m$^2$), and shortwave radiation (SWRad) (W/m$^2$) at multiple locations (longitude and latitude).

We extract both the weather and load data for the intersection period of 2011-2021 from these two datasets to be used for the ML-based load prediction in this paper. The load data is the aggregated load over all weather zones, which is the ERCOT system-wide total load. We extract the climate data at 8 representative locations corresponding to the 8 ERCOT weather zones respectively. To be noticed, the ERCOT load data and NLDAS climate data have different time zones: the central standard time (CST) and universal time coordinated (UTC) respectively. To be consistent, the climate data we extracted are converted into CST.

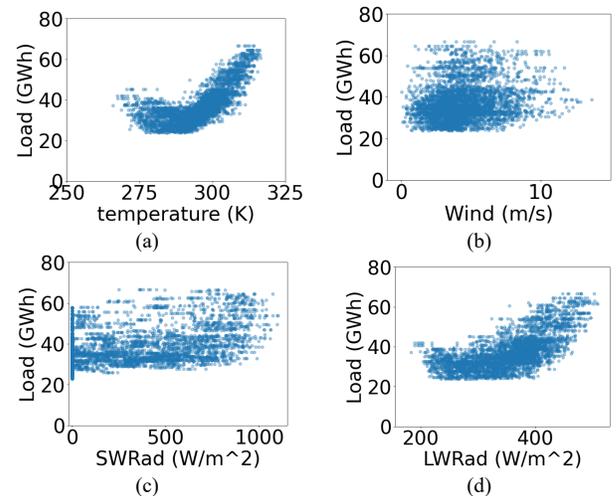

Fig. 3. Load plotted against weather conditions: (a) temperature, (b) wind speed, (c) short-wave radiation, and (d) long-wave radiation.

Fig. 3 shows the relation between several weather features and electrical load. It is clear that some features such as temperature are more closely related to the load than other features, such as shortwave radiation.

*B. Data Preprocessing*

First, we estimate the wind speeds correlated to the wind farm production by taking the root sum square of meridional and azimuthal wind speeds. The combined wind speed will be used in this work. Because time series inputs are used, we created subarrays of length ($T1+T2$), each as a training sample with the first $T1$ hours as inputs for look-back window, and the last $T2$ hours as outputs in the prediction window, by making copies of all parts of the data for which a consecutive of ($T1+T2$)-hour sequence can be found. In this work, the look-back window of $T1$ is 6-hour while the look-ahead prediction window of $T2$ is 4-hour. We normalized the raw data to assist in training. The input data into the model are normalized, while the outputs of ML models are denormalized to obtain the actual load predictions. Lastly, we split the data into training, and validation, and testing datasets by proportions of 45%, 45%, and 10% respectively. The independence of the training, validation, and testing datasets help prevent overfitting ML models to a specific dataset.

## IV. CASE STUDIES

We used Python script for performing most of the data processing, model training, and evaluation. We used the Keras library for creating the FCNN, LSTM, and LRCN models, while the Scikit-Learn library is used to train the SVM model.

### A. Predictor Training

We trained and tested the aforementioned ML models and their variations with various architecture and features, as well as a naïve benchmark method. The benchmark simply uses most recent hour's load as the future loads.

Fig. 4(a)-(c) shows the predicted load against the actual load for three models: benchmark, SVM and LSTM, respectively. By comparing these plots, it is observed that the benchmark has the least accuracy with an MAPE of 8.087% and a $R^2$ of 0.744, while both SVM and LSTM achieved much better results. Fig. 4(d) shows the LSTM prediction error distribution of all samples, indicating the prediction error of most samples are less than 5%.

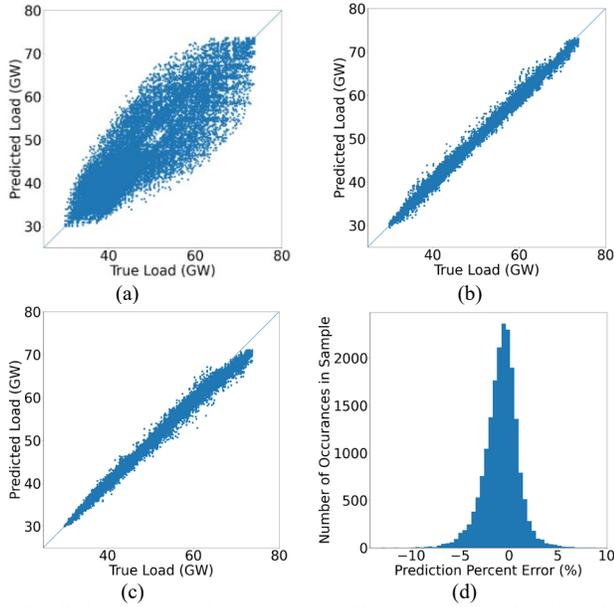

Fig. 4. Prediction model performance: (a) predicted load of benchmark against actual load; (b) predicted load of SVM against actual load; (c) predicted load of LSTM against actual load, and (d) prediction error distribution of LSTM.

Table I presents the prediction accuracies with different ML models. It shows FCNN has the least performance among all ML models tested. While both LSTM and LRCN can achieve very good performance, SVM has the best performance. This maybe because a separate SVM model is trained in order to predict the load of each successive hour. Because of the different method used to creating SVM models, we will primarily focus on comparing the results of the other models. The second best performing model is an LSTM model, with a MAPE of 1.336% and $R^2$ of 0.991, closely followed by an LRCN model with MAPE of 1.359% and $R^2$ of 0.991.

Table I Prediction accuracies with different ML models

| Model | SVM | FCNN | LSTM | LRCN |
|---|---|---|---|---|
| MAPE | 1.260% | 1.683% | 1.336% | 1.359% |
| $R^2$ | 0.994 | 0.986 | 0.991 | 0.991 |

### B. Sensitivity on Various Feature Combinations

Table II shows a select set of results from LSTM models of the same architectures trained on different features. LSTM model 1 was trained with only the previous *T1* hours of true electrical load as features. For models 2 through 5, only a single weather feature is included along with the load as input. Less rigorously, the results of these models largely support that the model in which there is a more discernible pattern between a weather feature and the load will produce more accurate load predictions, although there appears to be an exception for shortwave radiation. We believe that this results from the interaction of the load and shortwave radiation from previous hours. The fluctuation in shortwave radiation accurately forecasts the change in electrical load while the previous value of electrical load will better predict the magnitude of electrical load. As opposed to shortwave radiation on its own. LSTM models 1-2 and 6-10 show the prediction accuracy improves with more features to a certain extent before it decreases possibly due to overfitting. Although deep learning models are typically capable of handling large dimensions of input with redundancies, our study finds that using Load, Hour, Month, and Temperature (model 7), LSTM can achieve the highest accuracy. LSTM has procured a greater amount of "useful" information in model 7 than 8, indicating that either the model with the current architecture is incapable of handling more information, or that some features may be generally useless and distract the model.

Table II Prediction accuracies with LSTM models under different features

| Model No. | Input Features | MAPE | $R^2$ |
|---|---|---|---|
| 1 | Loads only | 2.288% | 0.973 |
| 2 | Load, Temp | 1.982% | 0.980 |
| 3 | Load, SWrad | 1.414% | 0.991 |
| 4 | Load, LWrad | 2.133% | 0.976 |
| 5 | Load, Wind | 2.051% | 0.977 |
| 6 | Load, Hour, Temp | 1.892% | 0.980 |
| 7 | Load, Hour, Month, Temp | 1.265% | 0.992 |
| 8 | Load, Hour, Month, Temp, SWrad | 1.388% | 0.989 |
| 9 | Load, Hour, Month, Temp, SWrad, Wind | 1.299% | 0.991 |
| 10 | Load, Hour, Month, Temp, SWrad, LWrad, Wind | 1.384% | 0.992 |
| 11 | Load, Hour, Day, Month, SWrad, Temp, LWrad, Wind | 1.411% | 0.991 |

Table III Prediction accuracies with larger LSTM model

| Model No. | Input Features | MAPE | $R^2$ |
|---|---|---|---|
| 1 | Loads only | 2.063% | 0.977 |
| 2 | Load, Temp | 1.953% | 0.983 |
| 3 | Load, SWrad | 1.434% | 0.990 |
| 4 | Load, LWrad | 2.135% | 0.978 |
| 5 | Load, Wind | 2.085% | 0.979 |
| 6 | Load, Hour, Temp | 1.880% | 0.982 |
| 7 | Load, Hour, Month, Temp | 1.454% | 0.991 |
| 8 | Load, Hour, Month, Temp, SWrad | 1.332% | 0.991 |
| 9 | Load, Hour, Month, Temp, SWrad, Wind | 1.324% | 0.991 |
| 10 | Load, Hour, Month, Temp, SWrad, LWrad, Wind | 1.654% | 0.988 |
| 11 | Load, Hour, DoW, Month, SWrad, Temp, LWrad, Wind | 1.282% | 0.992 |

Table III presents the prediction accuracies of larger LSTM models with neurons doubled in each hidden layer than Table II. In the table, it is also observed that not all feature additions are beneficial to the model. This indicates that using all information in training models may not yield the best results, justifying a need of feature selection.

The strong ability to predict time-series data may have made some features, such as the hour and temperature, less useful in LSTM than they may be in other models. We tested the other models in consideration, which demonstrated generally similar patterns; however, there are some notable differences. Table IV shows the testing results of the FCNN model. As more features are added, trends of increasing then decreasing accuracies persists. Comparatively speaking, LSTM models were significantly more accurate than FCNN models, possibly owing to the advantage of LSTM in capturing temporal correlations.

Table IV Prediction accuracies with FCNN under different features

| Model No. | Model | MAPE | $R^2$ |
|---|---|---|---|
| 1 | Loads only | 3.864% | 0.926 |
| 2 | Load, Temp | 2.531% | 0.962 |
| 3 | Load, SWrad | 1.634% | 0.988 |
| 4 | Load, LWrad | 2.986% | 0.948 |
| 5 | Load, Wind | 3.159% | 0.951 |
| 6 | Load, Hour, Temp | 2.050% | 0.980 |
| 7 | Load, Hour, Month, Temp | 1.386% | 0.991 |
| 8 | Load, Hour, Month, Temp, SWrad | 1.345% | 0.991 |
| 9 | Load, Hour, Month, Temp, SWrad, Wind | 1.368% | 0.991 |
| 10 | Load, Hour, Month, Temp, SWrad, LWrad, Wind | 1.435% | 0.991 |
| 11 | Load, Hour, DoW, Month, SWrad, Temp, LWrad, Wind | 1.440% | 0.990 |

## C. Ablation Studies

An ablation study was conducted to perform the sensitivity analysis of different climate information. The LSTM model was evaluated against identical models trained with various missing features. The models are evaluated and compared in Table V using MAPE, as well as accuracies under five different error thresholds. The model named 'ALL' means it uses all available features. The ablation studies shows that the influence of removing individual climate features such as wind information on the load prediction is quite obvious; after removing some key individual features such as LWRad, the accuracies became less than FCNN. However, there is an exception with the removal of temperature information, which seems to be largely predictable from other factors. The model of 'Time only" does not use any weather condition and can achieve similar prediction accuracy with FCNN but much less than the model 'ALL'.

We also repeated the training process of varying weather data for SVM, FCNN, and LRCN models and compared the outcomes of each model across the same weather feature inputs. These three models are trained with a similar set of architectures. Each is trained on the same training data and tested on the same test data. We observed similar observations with LSTM.

In this work, we aimed to justify the need for feature and model selection for short-term load forecasting. Case studies are focused on weather and time feature-trained ML models. To improve the model, we experimented with feature and model selection in downscaled situations and observed how the changes affect the accuracy of the model. We observed that there is significant overlap between time-series information for LSTM models including hour, temperature, and shortwave radiation intensity. As a result, the inclusion of several such features will not improve the model significantly compared to including just one such feature. Of these, shortwave radiation has the strongest effect on the prediction accuracy. Month is a significant predictor of load. The day-of-week, wind speed, and longwave radiation intensity are all weak predictors of load, with longwave radiation having the weakest association.

Table V Ablation study results of LSTM model

| Models | Accuracy under different error tolerances | | | | | MAPE |
|---|---|---|---|---|---|---|
| | 1% | 2% | 3% | 4% | 5% | |
| ALL | 46.8% | 76.6% | 91.1% | 96.4% | 98.5% | 1.376% |
| LWRad removed* | 31.6% | 60.4% | 81.5% | 92.2% | 96.6% | 1.881% |
| SWRad removed* | 41.1% | 71.4% | 88.1% | 94.9% | 97.8% | 1.542% |
| Wind removed* | 37.8% | 67.9% | 86.5% | 94.7% | 97.7% | 1.643% |
| Temperature removed* | 47.7% | 75.9% | 89.2% | 95.1% | 97.8% | 1.412% |
| Time only | 44.4% | 71.4% | 84.9% | 92.7% | 96.8% | 1.584% |
| FCNN | 42.5% | 71.4% | 86.6% | 93.7% | 97.1% | 1.579% |

*the weather feature that is removed in training, from all 8 weather zones.

## V. CONCLUSIONS

Accurate load forecasting is critical for efficient and reliable operations of the electric power system. This paper implemented multiple ML models for load forecasting and examined the effect of different input features particularly weather and time factors on the model accuracy. Various ML models were first constructed to compare model accuracies and inference capabilities. These ML models were then tested, after re-training, with varying input features to compare the effect of different feature combinations on the model accuracy. Lastly, results were compared to evaluate the effectiveness of including different features in the models.

The experimental results of model accuracies indicate that the incorporation of temporal data (hour in the day, day in the week, and month in the year) is useful in some sense. Short-term features such as the hour and temperature provide unnecessary redundancy, which may even decrease accuracy in models. Combinations of different scales of features generally improved model accuracy. LSTMs handled the time-series data with greater accuracy than FCNNs. Future work on this topic should explore how to efficiently test for features with overlapping information. Additionally, graph neural network that is uniquely designed to process spatial data may enhance the inference capabilities of ML models [25]-[26]; future work may examine ML models that combine both GNN layers and temporal layers.